%% file: main.tex
\documentclass[10pt,twocolumn,letterpaper]{article}

\usepackage{setspace}
\usepackage[accsupp]{axessibility}
\usepackage[pagenumbers]{cvpr} 

\input{preamble}
\usepackage{microtype}

\definecolor{cvprblue}{rgb}{0.21,0.49,0.74}
\usepackage[pagebackref,breaklinks,colorlinks,allcolors=cvprblue]{hyperref}
\usepackage{booktabs}

\def\paperID{48} 
\def\confName{CVPR}
\def\confYear{2026}

\title{Boxes2Pixels: Learning Defect Segmentation from Noisy SAM Masks}

\author{
Camile Lendering \quad
Erkut Akdag \quad
Egor Bondarev\\
AIMS Group, Department of Electrical Engineering, Eindhoven University of Technology\\
{\tt\small \{c.r.lendering, e.akdag, e.bondarev\}@tue.nl}
}

\begin{document}
\maketitle
\input{sec/0_abstract}    
\input{sec/1_intro}
\input{sec/2_related_works}
\input{sec/3_method}
\input{sec/4_dataset}
\input{sec/4_expirements}
\input{sec/5_conclusion}
{
    \small
    \bibliographystyle{ieeenat_fullname}
    \bibliography{main}
}


\end{document}

%% file: preamble.tex
\usepackage{booktabs}
\usepackage{bbm}

\usepackage{microtype}

\renewcommand{\paragraph}[1]{\vspace{.5em}\noindent\textbf{#1.}}

\usepackage{soul}
\setuldepth{foobar}

%% file: sec/0_abstract.tex
\begin{abstract}
Accurate defect segmentation is critical for industrial inspection, yet dense pixel-level annotations are rarely available. A common workaround is to convert inexpensive bounding boxes into pseudo-masks using foundation segmentation models such as the Segment Anything Model (SAM). However, these pseudo-labels are systematically noisy on industrial surfaces, often hallucinating background structure while missing sparse defects.

To address this limitation, a noise-robust box-to-pixel distillation framework, Boxes2Pixels, is proposed that treats SAM as a noisy teacher rather than a source of ground-truth supervision. Bounding boxes are converted into pseudo-masks offline by SAM, and a compact student is trained with (i) a hierarchical decoder over frozen DINOv2 features for semantic stability, (ii) an auxiliary binary localization head to decouple sparse foreground discovery from class prediction, and (iii) a one-sided online self-correction mechanism that relaxes background supervision when the student is confident, targeting teacher false negatives.

On a manually annotated wind turbine inspection benchmark, the proposed Boxes2Pixels improves anomaly mIoU by +6.97 and binary IoU by +9.71 over the strongest baseline trained under identical weak supervision. Moreover, online self-correction increases the binary recall by +18.56, while the model employs 80\% fewer trainable parameters. Code is available at \url{https://github.com/CLendering/Boxes2Pixels}.
\end{abstract}

%% file: sec/1_intro.tex
\section{Introduction}
\label{sec:intro}

Visual inspection of critical infrastructure, such as wind turbine blades, is essential for predictive maintenance and failure prevention.
Although automated inspection systems promise substantial reductions in operational cost and downtime, their deployment in industrial settings remains limited due to the scarcity of high-quality pixel-level annotations.

Bounding-box supervision offers a practical alternative.
Modern object detection models (e.g., YOLO~\cite{redmon2016you} and SSD~\cite{liu2016ssd}) can be trained efficiently from box annotations and are already widely adopted in industrial inspection pipelines~\cite{foster2022drone,zhanfang2025enhancing}.
However, downstream damage assessment and maintenance planning often require fine-grained segmentation.
Distinguishing superficial surface contamination from structural defects, such as cracks or delamination, depends on accurate spatial extent and local texture information, which cannot be reliably recovered from coarse object localization alone.
Moreover, pixel-level annotation is labor-intensive, costly, and requires domain expertise thereby severely limiting scalability.

\begin{figure}
    \centering
    \includegraphics[width=\linewidth]{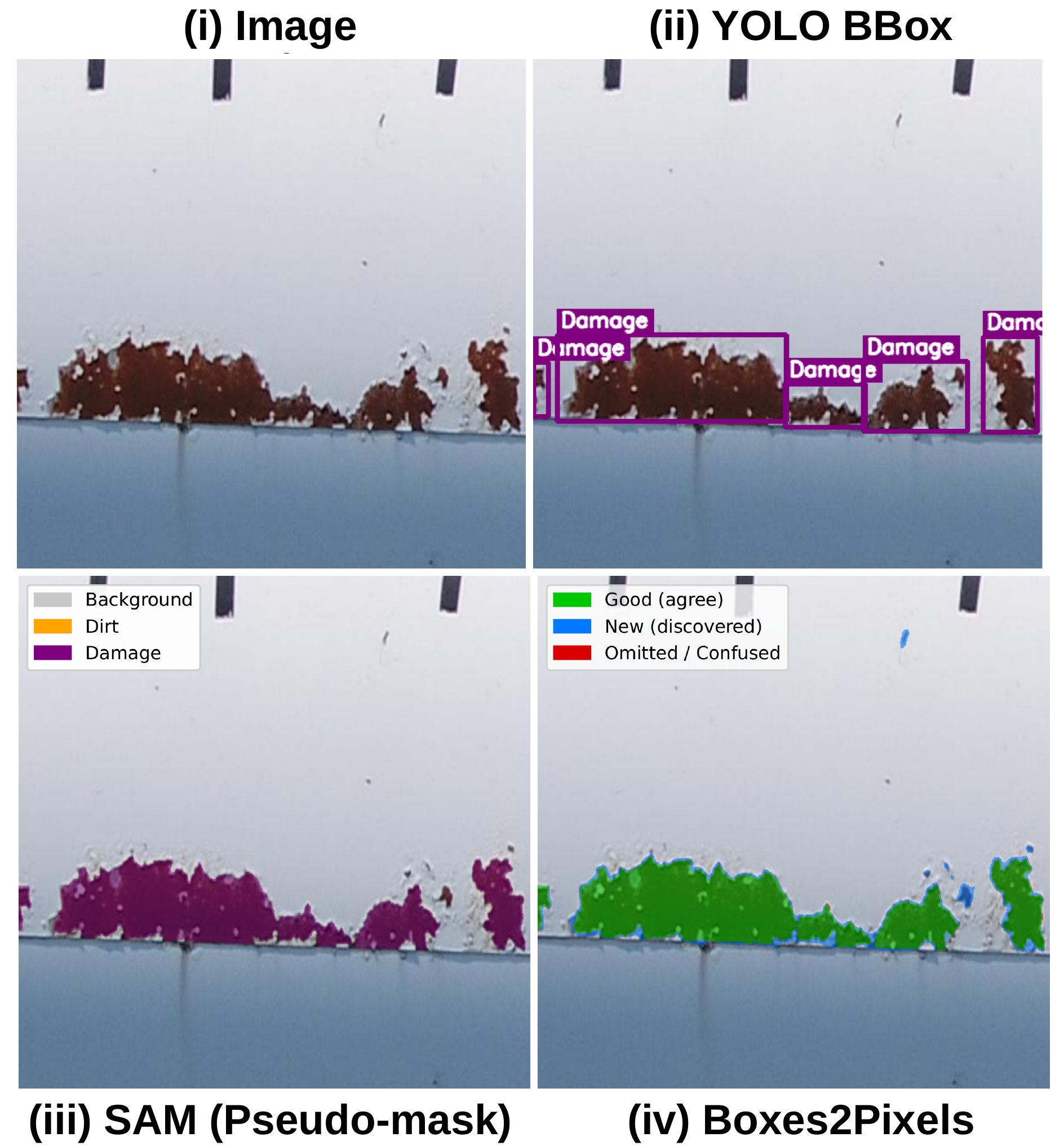}
    \caption{Overview of the proposed framework, Boxes2Pixels, for sparse defect segmentation.
    From left to right, top to bottom: (i) an input wind turbine image, (ii) legacy YOLO bounding boxes used as coarse supervision, (iii) SAM-generated pseudo-masks exhibiting both spurious regions and missed defects, and (iv) prediction from Boxes2Pixels.
    Colors indicate correct detections (green), defects recovered by the student but missed by the teacher (blue), and omitted or erroneous regions (red).}
    \label{fig:intro}
\end{figure}

Recent Vision Foundation Models (VFMs), e.g., SAM~\cite{kirillov2023segment} provides a practical pathway from box-level supervision and pixel-level segmentation.
Given simple prompts, such as bounding boxes, SAM can generate dense masks in a zero-shot manner.
However, applying SAM directly to industrial inspection remains impractical. The large SAM encoder incurs substantial computational overhead, and more critically, its predictions on industrial imagery exhibit~\emph{systematic errors}, including over-segmentation of background structures (e.g., shadows and texture patterns) and missed detection of thin or low-contrast defects.
Therefore, SAM-generated pseudo-masks are noisy and spatially inconsistent, limiting their usefulness as direct supervision signals.

This work proposes a \emph{data-centric distillation} framework, Boxes2Pixels, that leverages inexpensive bounding box annotations to train a reliable pixel-level defect segmentation model (Figure~\ref{fig:intro}).
During training, SAM is used to convert bounding boxes into initial pseudo-masks, which act as weak supervision for a lightweight student network.
Crucially, SAM is not required at inference time: once trained, the student performs dense defect segmentation directly from the input image.

A key challenge in this setting is robustness to pseudo-label noise.
In the experiments, conventional segmentation architectures, such as U-Net~\cite{ronneberger2015u} and SegFormer~\cite{xie2021segformer}, tend to overfit the systematic errors in SAM-generated masks, thereby inheriting the teacher’s failure modes.
To address this limitation, a hierarchical segmentation architecture is introduced, built upon frozen DINOv2-Small features~\cite{oquab2023dinov2}.
The semantic stability of self-supervised transformer representations reduces sensitivity to high-frequency artifacts and spurious boundaries, while a lightweight detail branch preserves local structures relevant for fine-grained defect localization.

Additionally, a~\emph{self-correcting loss} is introduced to relax the assumption of teacher correctness.
Instead of enforcing strict consistency with pseudo-labels, the proposed loss allows the student to override unreliable supervision when predicting defect classes with high confidence.
This asymmetric correction is motivated by a common failure mode in industrial settings: damage annotations are often incomplete, and subtle or sparsely distributed defects may be omitted from both bounding-box supervision and foundation-model outputs.
By allowing the student to deviate from background pseudo-labels when confident defect evidence is present, the model can recover defects missed by the teacher while still benefiting from reliable supervision.
To summarize, the main contributions of this work are as follows:

\begin{itemize}
    \item A data-centric distillation pipeline is introduced to learn pixel-level defect segmentation from bounding box annotations using noisy foundation-model pseudo-labels.
    \item A hierarchical student architecture is proposed based on frozen DINOv2 ViT-S/14 features, combined with a one-sided self-correcting loss to enable robust learning under systematic pseudo-label noise.
    \item Extensive evaluation is conducted on a wind turbine blade defect dataset, demonstrating improved robustness and recall compared to standard segmentation baselines trained under identical weak supervision.
\end{itemize}

%% file: sec/2_related_works.tex
\section{Related Work}
\label{sec:related}

\subsection{Weakly- and box-supervised segmentation} 
Early weakly-supervised semantic segmentation (WSSS) methods rely on class activation maps (CAMs)~\cite{zhou2016learning,selvaraju2017grad,ahn2018learning,wang2020self}, which typically yield coarse localization and fail to capture fine-grained defect topology.
More recent box-supervised segmentation approaches, including BoxSup~\cite{dai2015boxsup}, BoxInst~\cite{tian2021boxinst} and DiscoBox~\cite{lan2021discobox}, eliminate pixel-level supervision by combining projection constraints with pairwise pixel affinities.
While effective on natural-image benchmarks (e.g., COCO~\cite{lin2014microsoft}), these methods implicitly assume that the target occupies a substantial portion of the bounding box and exhibits relatively compact spatial structure.

In industrial inspection, these assumptions are often violated in two key ways.
First, many defects are sparse, elongated, or high-frequency (e.g., hairline cracks, scattered pitting), which weakens affinity propagation and leads over-smoothing.
Second, legacy bounding boxes are frequently loose, with defects occupying only a small fraction of the annotated region, which causes projection-based objectives to be dominated by background pixels.
As a result, box-supervised models tend to collapse to segmenting the entire bounding box rather than isolating the sparse anomalies.

Our approach mitigates this limitation by leveraging a foundation segmentation model to provide explicit pixel-level priors that are independent of defect fill-rate within the bounding box.

\subsection{Foundation models for industrial inspection} 
Vision foundation models pretrained on large-scale data, such as CLIP~\cite{radford2021learning}, MAE~\cite{he2022masked}, DINO~\cite{caron2021emerging}, and DINOv2~\cite{oquab2023dinov2}, have enabled advancements in industrial inspection pipelines through unsupervised anomaly localization and supervised distillation.
Unsupervised methods, such as PatchCore~\cite{roth2022towards} and WinCLIP~\cite{jeong2023winclip}, localize deviations from a learned normal manifold, but do not provide defect-type segmentation required for downstream inspection, reporting, or repair planning.
Moreover, their performance depends on the assumption that normal appearance variation is comprehensively represented during training. While effective for anomaly localization, these approaches do not address fine-grained defect segmentation.

In contrast, segmentation-oriented VFMs, most notably SAM~\cite{kirillov2023segment}, enable zero-shot mask prediction from sparse prompts and have been applied to industrial imagery.
However, their computational cost limits large-scale deployment, motivating distillation-based variants, such as MobileSAM~\cite{zhang2023faster}, FastSAM~\cite{zhao2023fast} and EfficientViT-SAM~\cite{zhang2024efficientvit}, which train lightweight student models to approximate SAM outputs. 

Recent analyses of weakly-supervised semantic segmentation report a mismatch between improvements in pseudo-mask quality and gains in final segmentation accuracy. This suggests that downstream training dynamics play a critical role beyond mask generation alone~\cite{chen2025weakly, rong2023boundary}.
This observation motivates approaches that focus on~\emph{how pseudo-labels are utilized during training}, rather than solely on~\emph{how they are generated}.
In industrial settings, pseudo-label quality varies widely due to coarse box supervision, complex surface textures, illumination changes, and background clutter.

Accordingly, our method treats pseudo-masks as noisy supervision and explicitly models their unreliability during optimization.
To stabilize learning under such noise, the student leverages self-supervised representations from DINOv2~\cite{oquab2023dinov2}, which provide strong semantic structure and dense correspondence across appearance variations~\cite{hamilton2022unsupervised}.

\subsection{Robust learning with noisy labels}

Learning under label noise has been extensively studied, with methods ranging from risk correction~\cite{natarajan2013learning,patrini2017making} to agreement-based filtering and small-loss selection~\cite{han2018co,li2020dividemix}. 
In classification, selective training strategies like Sel-CL~\cite{li2022selective} reduce the impact of corrupted samples by prioritizing high-confidence predictions and contrastive consistency.

In segmentation and foundation-model distillation, noise is often manifested as structured or systematic errors rather than purely random corruption. Approaches that model annotator confusion~\cite{tanno2019learning} or prevent early memorization of noisy labels~\cite{xia2020robust} demonstrate that dense prediction tasks are particularly sensitive to spatially correlated label noise.
In practice, such noise includes hallucinated foreground regions (e.g., shadows misclassified as defects) and omissions, where thin or low-contrast defects are missed entirely.

Recent segmentation approaches address noisy pseudo-labels by estimating spatial reliability and suppressing supervision in uncertain regions, using entropy, predictive variance, or teacher–student disagreement.
Foundational work on uncertainty-aware loss weighting is introduced by Kendall and Gal~\cite{kendall2017uncertainties}, and later extended to dense prediction through uncertainty-aware self-ensembling and cross-teaching strategies~\cite{yu2019uncertainty,tarvainen2017mean,zhao2024sam}.

Our work emphasizes \emph{controlled self-correction}, in contrast to most existing strategies that focus on \emph{noise rejection}, i.e., down-weighting or ignoring regions where the teacher is likely incorrect. 
When a pseudo-label marks background but the student confidently predicts a defect, supervision is selectively relaxed, enabling correction of likely false negatives.
This asymmetric design targets missed sparse defects while maintaining safeguards against uncontrolled drift.

%% file: sec/3_method.tex
\section{Method}
\label{sec:method}

Weakly supervised defect segmentation is addressed using annotated \emph{bounding boxes} provided as ground-truth annotations in the standard YOLO format, consisting of a class identifier and normalized center and size parameters.
No object detector is trained or used in this work; YOLO serves solely as an annotation representation.
Pixel-level pseudo-masks are generated offline by prompting SAM~\cite{kirillov2023segment} with the annotated bounding boxes. 
On industrial imagery, such pseudo-labels exhibit systematic errors, most notably false negatives for thin or low-contrast defects. 
To address this limitation, a hierarchical student network is trained with objectives explicitly designed to recover missed defects while remaining robust to pseudo-label noise.

\begin{figure*}[t]
    \centering
    \includegraphics[width=\textwidth]{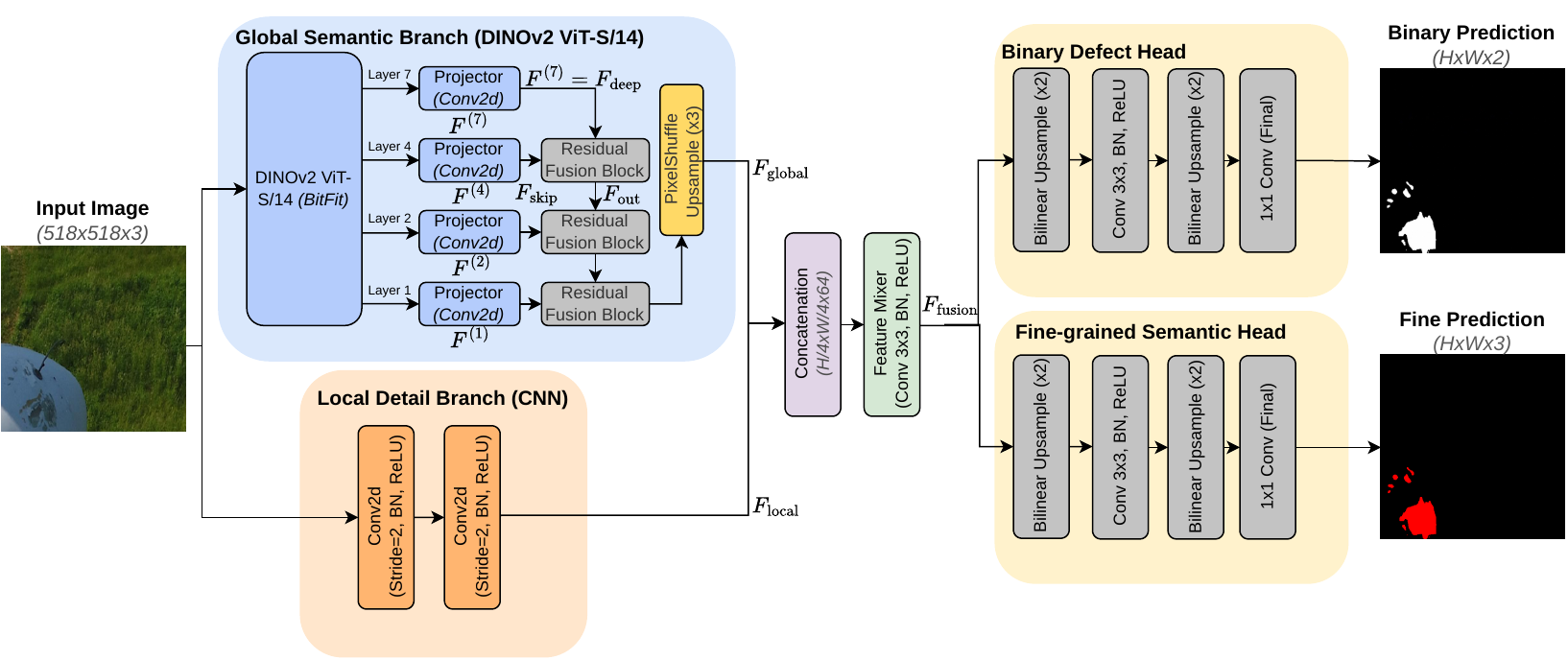}
    \caption{
Overview of the hierarchical student architecture.
The global semantic branch employs DINOv2 ViT-S/14 with BitFit adaptation to extract intermediate features from layers $\{7,4,2,1\}$.
Projected features $F^{(l)}$ are fused top-down through residual fusion blocks to produce $F_{\text{global}}$.
In parallel, a lightweight CNN detail branch extracts local features at the same resolution, yielding $F_{\text{local}}$.
Both branches are concatenated and processed by a feature mixer to obtain $F_{\text{fusion}}$, followed by a Binary Defect Head and a Fine-grained Semantic Head for $(K{+}1)$-class segmentation.
Training relies on offline SAM pseudo-masks generated from bounding boxes with objectives designed to mitigate systematic teacher errors.
    }
    \label{fig:architecture}
\end{figure*}

\subsection{Problem Formulation}
Pixel-level defect segmentation is studied under weak supervision, where only bounding box annotations are available and pixel-level ground-truth masks are not provided during training. 
Let $\mathcal{D}=\{(x_i,\mathcal{B}_i)\}_{i=1}^N$ denote a dataset of industrial images $x_i\in\mathbb{R}^{H\times W\times 3}$ and associated bounding box annotations

\begin{equation}
    \mathcal{B}_i=\{(b_{i,m},k_{i,m})\}_{m=1}^{M_i},
\end{equation}

where $M_i$ is the number of annotated bounding boxes in image $x_i$, and each bounding box $b_{i,m}$ is associated with a defect class label $k_{i,m}\in\{1,\dots,K\}$ (background corresponds to class $0$).

The objective is to learn a segmentation model

\begin{equation}
    S_\theta : x \mapsto \hat{y}\in[0,1]^{H\times W\times (K+1)},
\end{equation}

that predicts a per-pixel distribution over background and $K$ defect classes. 
Training relies on pseudo-labels derived from a foundation model, while the student is optimized to remain robust to systematic teacher errors.

\subsection{Teacher Pseudo-Labels from Bounding Boxes}
Direct supervision from bounding boxes yields ambiguous pixel assignments. 
Pseudo-labels are therefore generated by the Segment Anything Model (SAM). 
For each image $x_i$ and annotated bounding box $b_{i,m}\in\mathcal{B}_i$, the bounding box is provided as a prompt (without point or mask guidance), and SAM produces a single binary mask:
\begin{equation}
    \tilde{y}_{i}^{(m)} = \mathrm{SAM}(x_i, b_{i,m}).
\end{equation}
The pseudo-masks are generated offline and stored to ensure efficient student training.

To construct a multi-class training target, the per-bounding-box SAM masks are rasterized into a label image $\tilde{y}\in\{0,\dots,K\}^{H\times W}$ by assigning the class label $k_{i,m}$ to all pixels inside the predicted mask $\tilde{y}_{i}^{(m)}$. 
In the two-class setting (Dirt, Damage), YOLO class identifiers $\{0,1\}$ are mapped to pixel labels $\{1,2\}$, with background assigned to $0$.

Despite its strong zero-shot capability, SAM exhibits characteristic failure modes on industrial imagery, including false positives due to shadows, stains, and background texture, as well as false negatives where thin cracks or low-contrast defects are missed despite being enclosed by the bounding box. Therefore, the student architecture and training objective are designed to mitigate both error types.

\subsection{Hierarchical Student Architecture}
A hierarchical student architecture is constructed to combine stable global semantics with high-frequency spatial detail. 
Given an input RGB image $x\in\mathbb{R}^{H\times W\times 3}$ (resized to $518\times518$ during training), the network produces two outputs: 
(i) a binary defect map and (ii) a fine-grained $(K{+}1)$-class segmentation map at full image resolution. 
As illustrated in Figure~\ref{fig:architecture}, the architecture consists of a global semantic branch based on DINOv2 and a lightweight local detail branch, whose features are fused prior to prediction.

\textbf{Global Semantic Branch.}
DINOv2 ViT-S/14~\cite{oquab2023dinov2} is utilized as the semantic feature extractor.
All pretrained weights are frozen, while only bias and normalization parameters are optimized by BitFit-style adaptation~\cite{zaken2022bitfit}. BitFit updates only bias terms and normalization layers while keeping all other weights fixed, enabling lightweight domain adaptation and reducing overfitting under noisy pseudo-label supervision.

As shown in Figure~\ref{fig:architecture}, intermediate transformer activations are extracted from layers 
$L=\{1,2,4,7\}$.
Each layer output is reshaped into a 2D feature map and projected to a common decoder width via a $1\times1$ convolution, forming a multi-level feature hierarchy.

The multi-level transformer features are fused in a top-down manner.
Let $F^{(l)}$ denote the projected feature map from layer $l \in L$.
Fusion proceeds from the deepest to the shallowest layer.
The deepest feature $F^{(7)}$ initializes $F_{\text{deep}}$.
For each subsequent residual fusion block, $F_{\text{deep}}$ is fused with the next shallower lateral feature $F_{\text{skip}} = F^{(l)}$ via
\begin{equation}
    F_{\text{out}}
    =
    \sigma\!\left(
    \phi(F_{\text{deep}} + F_{\text{skip}})
    + (F_{\text{deep}} + F_{\text{skip}})
    \right),
\end{equation}
where $\phi(\cdot)$ is a lightweight Conv--BN--ReLU--Conv--BN module and $\sigma$ is ReLU. 
The resulting $F_{\text{out}}$ becomes the new $F_{\text{deep}}$ for the next fusion stage. 
After the final fusion step, the feature map
is progressively upsampled with learned PixelShuffle blocks to a quarter-resolution grid ($H/4\times W/4$), ensuring spatial consistency with the local detail branch for subsequent fusion. The final feature map of the global semantic branch is denoted as $F_{\text{global}}$. 

\textbf{Local Detail Branch.}
Transformer features are spatially coarse-grained (DINOv2 operates at patch resolution), which can limit sensitivity to thin or high-frequency defect structures. 
To complement global semantics, a lightweight CNN branch extracts finer-grained spatial features directly from the RGB input. 
Two stride-2 Conv--BN--ReLU blocks are employed to reduce the input resolution from $H\times W$ to $H/4\times W/4$, matching the spatial resolution of $F_{\text{global}}$. 
This design preserves higher spatial granularity than the ViT patch representation while maintaining computational efficiency. The output of the local branch is denoted as

\begin{equation}
    F_{\text{local}}(x)\in\mathbb{R}^{\frac{H}{4}\times \frac{W}{4}\times c_d},
\end{equation}

where $c_d$ is the channel dimension of the detail feature map. $F_{\text{local}}$ represents the final output of the local detail branch.

\textbf{Feature Fusion and Prediction Heads.}
The quarter-resolution feature maps $F_{\text{global}}$ and $F_{\text{local}}$ are concatenated along the channel dimension and processed by a $3\times3$ Conv--BN--ReLU Feature Mixer block to produce a unified representation $F_{\text{fusion}}$. 
The mixer learns to balance semantic consistency from $F_{\text{global}}$ with fine spatial cues from $F_{\text{local}}$. Two prediction heads operate on $F_{\text{fusion}}$ as follows.

\textit{Binary Defect Head.}
The binary head predicts foreground (defect) versus background. 
It consists of two bilinear upsampling stages (each $\times2$), interleaved with a $3\times3$ Conv--BN--ReLU block, followed by a final $1\times1$ convolution producing $2$ output channels.

\textit{Fine-Grained Semantic Head.}
The fine-grained head predicts $(K{+}1)$ semantic classes. 
It shares the same architectural structure as the binary head but produces $(K{+}1)$ output channels via the final $1\times1$ convolution.

Both heads take the same fused feature map $F_{\text{fusion}}$ as input but optimize different objectives (structural localization vs. semantic discrimination). 
Their logits are upsampled to full image resolution ($H\times W$) to produce the final segmentation outputs, as illustrated on the right side of Figure~\ref{fig:architecture}.

\subsection{Training Objectives}
Training the hierarchical student architecture minimizes a composite objective combining a structural localization loss for the binary head and a semantic classification loss for the fine-grained head.

\subsubsection{Binary head loss function}
Defects typically occupy a small fraction of an image, causing region-based losses to be dominated by background pixels. Therefore, the binary head is supervised by an Asymmetric Dice loss~\cite{hashemi2018asymmetric}:
\begin{equation}
\label{eq:asym_dice_clean}
\mathcal{L}_{\text{bin}}
=
1 -
\frac{\langle p, g \rangle + \epsilon}
{\langle p, g \rangle
+ \beta \langle p, 1-g \rangle
+ \langle 1-p, g \rangle
+ \epsilon},
\end{equation}
where $\langle\cdot,\cdot\rangle$ denotes summation over spatial locations, 
$p\in[0,1]^{H\times W}$ is the predicted per-pixel defect probability, and the binary target is derived from the pseudo-mask as $g_i=\mathbbm{1}[\tilde{y}_i>0]$.
The factor $\beta\in(0,1)$ down-weights false positives, biasing optimization toward recall; $\beta=0.4$ is used unless stated otherwise.
The constant $\epsilon>0$ ensures numerical stability by preventing division by zero in defect-free regions.

\subsubsection{Semantic loss function with online self-correction}
Direct supervision of the semantic head with the pseudo-labels $\tilde{y}$ risks propagating false negatives from the teacher. To address this issue, a one-sided online self-correction mechanism is introduced, which selectively relaxes background supervision when the model predicts a defect class with high confidence.
The corrected target is defined as
\begin{equation}
\label{eq:self_correction}
\tilde{y}_i^{\text{corr}} =
\begin{cases}
\arg\max_{c>0} p_{i,c}, &
\text{if } \tilde{y}_i = 0 \ \land\ \max_{c>0} p_{i,c} > \tau, \\
\tilde{y}_i, & \text{otherwise},
\end{cases}
\end{equation}
where $c=0$ denotes the background, $c>0$ correspond to defect classes, and $\tau\in(0,1)$ is a confidence threshold.
Thus, only background labels are eligible for correction, while annotated defect regions remain unchanged.
The correction is computed online per mini-batch and does not modify the training dataset.
A short warm-up phase can be employed during which $\tilde{y}^{\text{corr}}=\tilde{y}$ to stabilize early training.
The semantic head is optimized by 
\begin{equation}
\label{eq:semantic_loss}
\mathcal{L}_{\text{fine}} = \mathrm{CE}\!\left(\hat{y}_{\text{fine}}, \tilde{y}^{\text{corr}}\right),
\end{equation}
where $\mathrm{CE}(\cdot,\cdot)$ denotes class-weighted cross-entropy to mitigate the severe foreground--background imbalance. The final training objective is given by:
\begin{equation}
\label{eq:total_loss}
    \mathcal{L}_{\text{total}}
    =
    \lambda_{\text{bin}} \mathcal{L}_{\text{bin}}
    +
    \lambda_{\text{fine}} \mathcal{L}_{\text{fine}},
\end{equation}
where $\lambda_{\text{bin}}$ and $\lambda_{\text{fine}}$ balance structural localization and semantic discrimination.

\subsection{Training Protocol}
\label{ssec:training_protocol}
The student is trained end-to-end using AdamW~\cite{loshchilov2017decoupled} with a cosine learning-rate schedule.
The initial learning rate is set to $5\times10^{-4}$ with a weight decay of $1\times10^{-2}$.
Gradients are clipped to a maximum $\ell_2$ norm of $1.0$ to improve stability under noisy pseudo-label supervision.

To reduce optimization variance, an exponential moving average (EMA) of the student parameters is maintained during training~\cite{polyak1992acceleration}. Specifically, a second set of weights is updated as a running average of the current model parameters at each iteration with decay $0.999$. This EMA model is not a separate architecture, but a smoothed version of the student network, and its weights are used for validation and inference.

The binary and semantic losses are combined with equal weighting,
$\lambda_{\text{bin}}=\lambda_{\text{fine}}=0.5$,
which balances structural localization and semantic discrimination without introducing additional bias toward either objective. 
In practice, both losses operate at comparable scales after normalization, and equal weighting was found to provide stable convergence.

For the asymmetric Dice loss, the false-positive weighting factor is set to $\beta=0.4$. 
Following~\cite{hashemi2018asymmetric}, $\beta<1$ reduces the penalty on false positives and implicitly biases optimization toward recall, which is desirable in industrial inspection, where missed defects are more critical than slight over-segmentation.

Online self-correction is enabled with a high confidence threshold $\tau=0.9$. 
Such conservative thresholds are commonly adopted in pseudo-labeling and teacher–student frameworks to ensure that only highly confident predictions override existing supervision. 
Correction is disabled during an initial warm-up phase to prevent unstable updates.

%% file: sec/4_dataset.tex
\section{Dataset}
\label{sec:dataset}

A publicly available wind turbine inspection dataset is employed in this work.
The images originate from the \textit{DTU – Drone Inspection Images of Wind Turbine} dataset~\cite{shihavuddin2018dtu}, which was later released in a YOLO-annotated format by Foster et al.~\cite{foster2022drone} and made publicly available via Kaggle.
The dataset contains approximately 13,000 RGB images of size $586\times371$ captured in real-world outdoor environments.

Annotations are provided exclusively as bounding boxes in YOLO format for two defect categories: \emph{damage} and \emph{surface dirt}.
No pixel-level segmentation masks are available during training.
All models are, therefore, trained using bounding box supervision~\emph{only}, without access to pixel masks or auxiliary annotations.

The dataset exhibits substantial class imbalance.
Less than $25\%$ of images contain annotated defects, and when present, defect pixels typically occupy less than $5\%$ of the image area.
In addition, image quality varies significantly due to motion blur, illumination changes, shadows, and strong surface textures, making boundary-accurate defect segmentation particularly challenging and prone to systematic pseudo-label noise.

%% file: sec/4_expirements.tex
\section{Experiments and Results}
\label{sec:experiments}

The proposed box-to-pixel training framework is evaluated on real-world industrial imagery with a focus on robustness under noisy pseudo-label supervision.
A key practical challenge is \emph{incomplete box supervision} in the provided annotations: subtle defects are sometimes missing bounding boxes, and even when boxes exist, SAM may omit thin or low-contrast regions within them. Accordingly, the experiments address three questions:

\begin{enumerate} [label=(\roman*)]
    \item Does the proposed hierarchical student improve segmentation quality compared to standard architectures trained under \textit{identical} weak supervision?
    \item Which architectural and loss components contribute to robustness under noisy pseudo-labels?
    \item Does online self-correction enable recovery of defects missed by the teacher or absent from box annotations?
\end{enumerate}

\subsection{Experimental Setup}
\label{ssec:setup}

All experiments are conducted on the wind turbine inspection dataset described in Sec.~\ref{sec:dataset}.
The original dataset provides only bounding box annotations and does not contain pixel-level segmentation masks.

A realistic weakly supervised protocol is adopted.
All models are trained using only bounding box annotations, which are converted into pseudo-masks using SAM.
No manually annotated pixel-level ground-truth masks are used during training.

To enable reliable quantitative evaluation, a held-out test split was manually annotated at pixel level to produce pixel-accurate ground-truth segmentation masks.
Validation performance is computed against pseudo-masks and is used solely for model selection and early stopping.
Test performance is computed against the manually annotated pixel-level ground-truth masks.

\subsection{Quantitative Results}
\label{ssec:quant_res}

Quantitative results on the manually annotated test set are reported in Table~\ref{tab:main_results}. 
Mean Intersection-over-Union (mIoU) denotes the average IoU computed over all $(K{+}1)$ classes, including background. 
In contrast, anomaly mIoU (mIoU$_{\text{anom}}$) averages IoU only over the defect classes (Damage and Dirt), excluding background.

Binary IoU (IoU$_{\text{bin}}$) evaluates foreground–background segmentation by collapsing all defect classes into a single foreground category. 
Anomaly F1 (F1$_{\text{anom}}$) measures the harmonic mean of precision and recall over defect pixels, providing a recall-sensitive assessment that is particularly relevant for detecting missed defects under severe class imbalance.

\begin{table}[ht]
\centering
\caption{Quantitative comparison on the manually annotated test split (evaluation against pixel-level ground truth). All models are trained with identical box-only supervision via SAM pseudo-masks; higher is better for all metrics.}
\label{tab:main_results}
\footnotesize
\setlength{\tabcolsep}{3.5pt}
\begin{tabular}{lcccc}
\toprule
Model & mIoU & mIoU$_{\text{anom}}$ & F1$_{\text{anom}}$ & IoU$_{\text{bin}}$ \\
\midrule
U-Net & 0.7057 & 0.5629 & 0.7036 & 0.5427 \\
DeepLabV3-B2 & 0.6867 & 0.5342 & 0.6955 & 0.5331 \\
SegFormer-B2 & 0.7231 & 0.5881 & 0.6939 & 0.5312 \\
\midrule
\textbf{Boxes2Pixels (ours)} & \textbf{0.7661} & \textbf{0.6523} & \textbf{0.7674} & \textbf{0.6226} \\
\bottomrule
\end{tabular}
\end{table}

As shown in Table~\ref{tab:main_results}, the proposed method Boxes2Pixels outperforms all baselines across the metrics under the same weak supervision.
Boxes2Pixels achieves mIoU$_{\text{anom}}=0.6523$, improving over the strongest baseline (SegFormer-B2, 0.5881) by $+0.0642$.
The gain in IoU$_{\text{bin}}$ from 0.5312 to 0.6226 ($+0.0914$) indicates substantially improved foreground localization, which is critical when subtle defects are easily missed.

While SegFormer-B2 is a strong general-purpose segmenter, it shows limited robustness to the pseudo-label noise in this box-only setting.
In contrast, the proposed student combines semantically stable DINOv2 features (via constrained BitFit adaptation), a dedicated detail branch for thin structures, and explicit binary localization. This combination reduces sensitivity to spurious teacher boundaries while improving the recovery of sparse defects.

\subsection{Ablation Study}
\label{ssec:ablation}

The contribution of individual components is evaluated through ablation studies on the validation split. Since this split is assessed against SAM pseudo-labels, the resulting metrics reflect alignment with the noisy teacher and they are used for relative comparison and checkpoint selection.
The impact of the local detail branch (CNN), the auxiliary binary head, and the self-correcting loss is explored in Table~\ref{tab:ablation_val}. 

\begin{table}[ht]
\centering
\caption{Ablation study on the validation split (evaluation against SAM pseudo-labels). Metrics reflect agreement with the noisy teacher and are used for comparison and checkpoint selection.}
\label{tab:ablation_val}
\footnotesize
\begin{tabular}{lcccc}
\toprule
Configuration & mIoU & mIoU$_{\text{anom}}$ & F1$_{\text{anom}}$ & IoU$_{\text{bin}}$ \\
\midrule
Boxes2Pixels & 0.6709 & 0.5093 & 0.6761 & 0.5107 \\
w/o Local Detail Branch & 0.6655 & 0.5013 & 0.6631 & 0.4960 \\
w/o Binary Head & 0.5868 & 0.3851 & 0.5693 & 0.3979 \\
w/o Self-Correction & 0.6679 & 0.5049 & 0.6622 & 0.4949 \\
\bottomrule
\end{tabular}
\end{table}

As seen in Table~\ref{tab:ablation_val}, removing the~\emph{binary head} results in the largest performance drop, with mIoU$_{\text{anom}}$ decreasing from 0.5093 to 0.3851.
This highlights the importance of explicit structural supervision for foreground localization, which decouples sparse defect discovery from fine-grained classification and mitigates overfitting to background-dominated pseudo-labels. Second,~\emph{the local detail branch} yields smaller but consistent improvement, supporting its role in preserving thin and high-frequency structures that are often blurred by transformer-based decoders.
Finally, self-correction yields only a small change on validation, since the pseudo-labels may not contain the missed defects it aims to recover; its effect is, therefore, evaluated on the manually annotated test set (Sec.~\ref{ssec:self_correction}).

\subsection{Impact of Self-Correction}
\label{ssec:self_correction}

To assess the effect of asymmetric online self-correction, two models trained under identical settings are compared, differing only in whether label correction is enabled.
Evaluation is conducted on the manually annotated test set, ensuring that improvements reflect genuine recovery of defects rather than improved agreement with the teacher.

\begin{table}[ht]
\centering
\caption{Impact of online self-correction on the manually annotated test split (evaluation against pixel-level ground truth). Two models differ only by enabling correction; Recall$_{\text{bin}}$ highlights missed-defect recovery.}
\label{tab:self_correction}
\footnotesize
\setlength{\tabcolsep}{2.5pt}
\begin{tabular}{lcccc}
\toprule
Method & mIoU$_{\text{anom}}$ & F1$_{\text{anom}}$ & IoU$_{\text{bin}}$ & Recall$_{\text{bin}}$ \\
\midrule
w/o Self-Correction & 0.5826 & 0.6889 & 0.5255 & 0.6195 \\
\textbf{w/ Self-Correction (Ours)} & \textbf{0.6523} & \textbf{0.7674} & \textbf{0.6226} & \textbf{0.8051} \\
\midrule
\textit{Improvement} & \textit{+0.0697} & \textit{+0.0785} & \textit{+0.0971} & \textit{\textbf{+0.1856}} \\
\bottomrule
\end{tabular}
\end{table}

Self-correction improves all reported metrics as illustrated in Table~\ref{tab:self_correction}.
The strongest effect is the increase in binary recall from 0.6195 to 0.8051 (\emph{+0.1856}), directly addressing the operational risk of missed defects.
This behavior is consistent with the underlying motivation: false negatives arise not only from SAM omissions within bounding boxes, but also from missing box annotations for subtle defects, both of which are treated as background during training.
By selectively relaxing background supervision, when the student exhibits high confidence, self-correction enables the recovery of sparse defects that would otherwise remain suppressed.

Importantly, these gains do not require increasing model capacity and are achieved without a disproportionate increase in false positives, indicating the correction mechanism remains conservative and does not induce semantic drift.

\subsection{Class-wise Analysis}
\label{ssec:classwise}

Table~\ref{tab:per_class} reports per-class IoU on the manually annotated test set.
Structural damage is safety-critical, whereas surface dirt is visually confounding but generally less severe.

\begin{table}[ht]
\centering
\caption{Per-class IoU on the manually annotated test set.}
\label{tab:per_class}
\footnotesize
\begin{tabular}{lccc}
\toprule
Model & Damage & Dirt & $\Delta_{\text{Damage}}$ \\
\midrule
DeepLabV3-B2 & 53.89 & 52.95 & -17.76 \\
U-Net & 61.01 & 51.57 & -10.64 \\
SegFormer-B2 & 71.48 & 46.13 & -0.17 \\
\midrule
\textbf{Boxes2Pixels (ours)} & \textbf{71.65} & \textbf{58.81} & -- \\
\bottomrule
\end{tabular}
\end{table}

The proposed method Boxes2Pixels matches the strongest baseline on the Damage class (71.65 vs.\ 71.48 for SegFormer-B2), while substantially improving Dirt (58.81 vs.\ 46.13).
This indicates that the proposed method does not trade off structural damage detection for contamination cues; instead, it improves performance across defect types, consistent with enhanced foreground localization and reduced sensitivity to spurious teacher boundaries.

\subsection{Model Efficiency}
\label{ssec:efficiency}

Table~\ref{tab:params} compares model complexity, highlighting the parameter efficiency of the proposed method Boxes2Pixels.
By freezing the DINOv2 backbone and adapting it with BitFit, the proposed approach significantly reduces the number of trainable parameters.

\begin{table}[ht]
\centering
\caption{Efficiency comparison. Boxes2Pixels achieves real-time throughput ($>160$ FPS) while requiring 80\% fewer trainable parameters for adaptation.}
\label{tab:params}
\resizebox{\columnwidth}{!}{
\footnotesize
\begin{tabular}{lccccc}
\toprule
Model & Total & Trainable & FLOPs & Latency & FPS \\
& Params (M) & Params (M) & (G) & (ms) & \\
\midrule
DeepLabV3-B2 & 20.5 & 20.5 & \textbf{18.92} & 7.99 & 125.1 \\
SegFormer-B2 & 27.3 & 27.3 & 58.97 & 13.33 & 75.0 \\
\textbf{Boxes2Pixels (ours)} & 27.6 & \textbf{5.6} & 43.37 & \textbf{6.20} & \textbf{161.4} \\
\bottomrule
\end{tabular}
}
\end{table}

Despite comparable total model capacity, the proposed method uses 5.6M trainable parameters, representing an 80\% reduction compared to fully trainable baselines.
This constrained adaptation limits the effective capacity exposed to noisy pseudo-labels and is attractive for industrial deployment, where rapid adaptation under limited compute and annotation budgets is required. Furthermore, despite higher theoretical FLOPs than DeepLabV3, Boxes2Pixels achieves the lowest inference latency (6.20 ms) and highest throughput (161.4 FPS) on an NVIDIA H100, proving its viability for real-time operation.

\subsection{Qualitative Results}
\label{ssec:qualitative}
\begin{figure}
    \centering
    \includegraphics[width=1.05\linewidth]{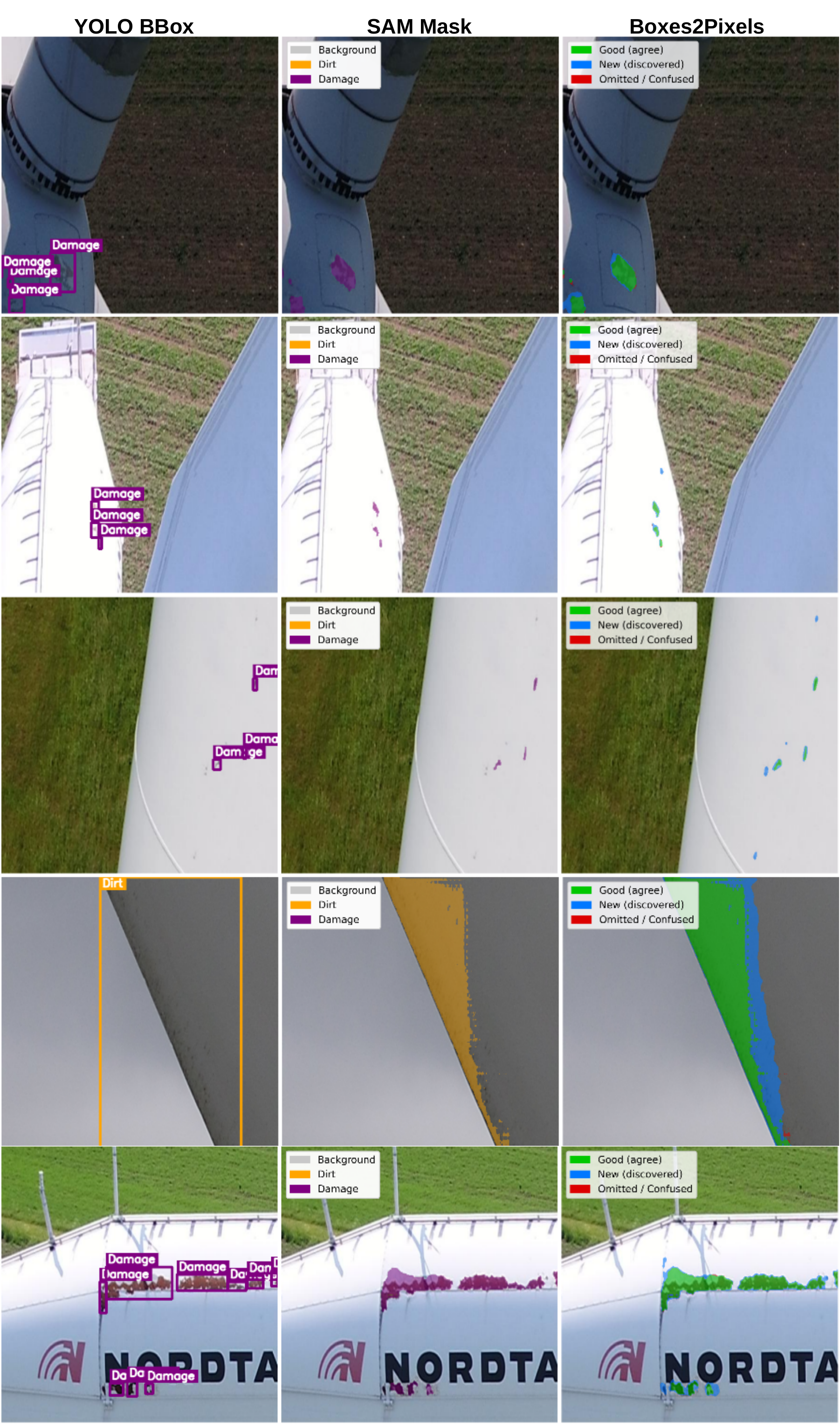}
    \caption{Qualitative comparison on wind turbine blade inspection images.
    From left to right: YOLO bounding boxes used for supervision, SAM-generated pseudo-masks, and predictions of the proposed Boxes2Pixels}
    \label{fig:qualitative}
\end{figure}

Figure~\ref{fig:qualitative} shows representative examples (YOLO boxes, SAM pseudo-masks, and student predictions).

The supervision exhibits systematic noise. 
SAM masks are often conservative, missing thin or low-contrast defects inside the box, while bounding boxes may omit subtle damage entirely. 
Conversely, box- and SAM-derived labels can introduce false positives by segmenting shadows, stains, or surface texture as defects.

The proposed student alleviates these issues through semantically stable features, a local detail branch, and noise-aware training. 
Compared to SAM, it recovers more complete defect regions, especially elongated cracks, while avoiding box-filling behavior on dirt. 
This aligns with the recall gains in Table~\ref{tab:self_correction} and reflects the effect of asymmetric self-correction, where confident predictions override unreliable pseudo-labels. Overall, the qualitative results support the quantitative improvements under box-only supervision.

%% file: sec/5_conclusion.tex
\section{Conclusion}

This work investigates whether reliable pixel-level defect segmentation can be learned from bounding-box supervision in real-world industrial inspection, where labels are incomplete and foundation-model pseudo-masks are systematically noisy.
Rather than treating foundation models as oracles, a data-centric distillation framework is introduced to explicitly address dominant error modes, including missed defects caused by SAM omissions or incomplete annotations.

Starting from bounding box prompts, SAM pseudo-masks are generated offline and used to train a compact hierarchical student with noise-aware objectives.
Frozen DINOv2 features provide semantic stability, while a binary localization head and an asymmetric online self-correction mechanism enable conservative recovery of missed defects without overriding teacher supervision.
On a manually annotated pixel-level turbine inspection benchmark, the proposed Boxes2Pixels outperforms standard weakly supervised baselines, with substantial gains in recall and anomaly IoU.

Beyond accuracy, the method is practical: BitFit-style adaptation reduces the number of trainable parameters by 80\%, enabling efficient domain adaptation under limited compute and annotation budgets.
Overall, the results suggest that robust utilization of pseudo-labels during training can be more impactful than refining pseudo-label generation alone.
Future work will explore transfer to additional industrial domains and uncertainty-aware self-correction.

\section*{Acknowledgements} 
\raggedright 
This work is supported by the ADVISOR ITEA 241007 project.

%% file: main.bib
@inproceedings{redmon2016you,
  title={You only look once: Unified, real-time object detection},
  author={Redmon, Joseph and Divvala, Santosh and Girshick, Ross and Farhadi, Ali},
  booktitle={Proceedings of the IEEE conference on computer vision and pattern recognition},
  pages={779--788},
  year={2016}
}

@inproceedings{kirillov2023segment,
  title={Segment anything},
  author={Kirillov, Alexander and Mintun, Eric and Ravi, Nikhila and Mao, Hanzi and Rolland, Chloe and Gustafson, Laura and Xiao, Tete and Whitehead, Spencer and Berg, Alexander C and Lo, Wan-Yen and others},
  booktitle={Proceedings of the IEEE/CVF international conference on computer vision},
  pages={4015--4026},
  year={2023}
}

@inproceedings{ronneberger2015u,
  title={U-net: Convolutional networks for biomedical image segmentation},
  author={Ronneberger, Olaf and Fischer, Philipp and Brox, Thomas},
  booktitle={International Conference on Medical image computing and computer-assisted intervention},
  pages={234--241},
  year={2015},
  organization={Springer}
}

@article{xie2021segformer,
  title={SegFormer: Simple and efficient design for semantic segmentation with transformers},
  author={Xie, Enze and Wang, Wenhai and Yu, Zhiding and Anandkumar, Anima and Alvarez, Jose M and Luo, Ping},
  journal={Advances in neural information processing systems},
  volume={34},
  pages={12077--12090},
  year={2021}
}

@article{oquab2023dinov2,
  title={Dinov2: Learning robust visual features without supervision},
  author={Oquab, Maxime and Darcet, Timoth{\'e}e and Moutakanni, Th{\'e}o and Vo, Huy and Szafraniec, Marc and Khalidov, Vasil and Fernandez, Pierre and Haziza, Daniel and Massa, Francisco and El-Nouby, Alaaeldin and others},
  journal={arXiv preprint arXiv:2304.07193},
  year={2023}
}

@inproceedings{zhou2016learning,
  title={Learning deep features for discriminative localization},
  author={Zhou, Bolei and Khosla, Aditya and Lapedriza, Agata and Oliva, Aude and Torralba, Antonio},
  booktitle={Proceedings of the IEEE conference on computer vision and pattern recognition},
  pages={2921--2929},
  year={2016}
}

@inproceedings{jeong2023winclip,
  title={Winclip: Zero-/few-shot anomaly classification and segmentation},
  author={Jeong, Jongheon and Zou, Yang and Kim, Taewan and Zhang, Dongqing and Ravichandran, Avinash and Dabeer, Onkar},
  booktitle={Proceedings of the IEEE/CVF Conference on Computer Vision and Pattern Recognition},
  pages={19606--19616},
  year={2023}
}

@article{zhang2023faster,
  title={Faster segment anything: Towards lightweight sam for mobile applications},
  author={Zhang, Chaoning and Han, Dongshen and Qiao, Yu and Kim, Jung Uk and Bae, Sung-Ho and Lee, Seungkyu and Hong, Choong Seon},
  journal={arXiv preprint arXiv:2306.14289},
  year={2023}
}

@inproceedings{zhang2024efficientvit,
  title={Efficientvit-sam: Accelerated segment anything model without performance loss},
  author={Zhang, Zhuoyang and Cai, Han and Han, Song},
  booktitle={Proceedings of the IEEE/CVF Conference on Computer Vision and Pattern Recognition},
  pages={7859--7863},
  year={2024}
}

@inproceedings{li2022selective,
  title={Selective-supervised contrastive learning with noisy labels},
  author={Li, Shikun and Xia, Xiaobo and Ge, Shiming and Liu, Tongliang},
  booktitle={Proceedings of the IEEE/CVF conference on computer vision and pattern recognition},
  pages={316--325},
  year={2022}
}

@inproceedings{zhao2024sam,
  title={Sam-Driven Weakly Supervised Nodule Segmentation with Uncertainty-Aware Cross Teaching},
  author={Zhao, Xingyue and Li, Peiqi and Luo, Xiangde and Yang, Meng and Chang, Shi and Li, Zhongyu},
  booktitle={2024 IEEE International Symposium on Biomedical Imaging (ISBI)},
  pages={1--5},
  year={2024},
  organization={IEEE}
}

@inproceedings{tian2021boxinst,
  title={Boxinst: High-performance instance segmentation with box annotations},
  author={Tian, Zhi and Shen, Chunhua and Wang, Xinlong and Chen, Hao},
  booktitle={Proceedings of the IEEE/CVF conference on computer vision and pattern recognition},
  pages={5443--5452},
  year={2021}
}

@inproceedings{lan2021discobox,
  title={Discobox: Weakly supervised instance segmentation and semantic correspondence from box supervision},
  author={Lan, Shiyi and Yu, Zhiding and Choy, Christopher and Radhakrishnan, Subhashree and Liu, Guilin and Zhu, Yuke and Davis, Larry S and Anandkumar, Anima},
  booktitle={Proceedings of the IEEE/CVF international conference on computer vision},
  pages={3406--3416},
  year={2021}
}

@inproceedings{roth2022towards,
  title={Towards total recall in industrial anomaly detection},
  author={Roth, Karsten and Pemula, Latha and Zepeda, Joaquin and Sch{\"o}lkopf, Bernhard and Brox, Thomas and Gehler, Peter},
  booktitle={Proceedings of the IEEE/CVF conference on computer vision and pattern recognition},
  pages={14318--14328},
  year={2022}
}

@article{hamilton2022unsupervised,
  title={Unsupervised semantic segmentation by distilling feature correspondences},
  author={Hamilton, Mark and Zhang, Zhoutong and Hariharan, Bharath and Snavely, Noah and Freeman, William T},
  journal={arXiv preprint arXiv:2203.08414},
  year={2022}
}

@inproceedings{zaken2022bitfit,
  title={Bitfit: Simple parameter-efficient fine-tuning for transformer-based masked language-models},
  author={Zaken, Elad Ben and Goldberg, Yoav and Ravfogel, Shauli},
  booktitle={Proceedings of the 60th Annual Meeting of the Association for Computational Linguistics (Volume 2: Short Papers)},
  pages={1--9},
  year={2022}
}

@article{hashemi2018asymmetric,
  title={Asymmetric loss functions and deep densely-connected networks for highly-imbalanced medical image segmentation: Application to multiple sclerosis lesion detection},
  author={Hashemi, Seyed Raein and Salehi, Seyed Sadegh Mohseni and Erdogmus, Deniz and Prabhu, Sanjay P and Warfield, Simon K and Gholipour, Ali},
  journal={IEEE Access},
  volume={7},
  pages={1721--1735},
  year={2018},
  publisher={IEEE}
}

@misc{shihavuddin2018dtu,
  title={DTU-Drone inspection images of wind turbine. Mendeley Data, V2},
  author={Shihavuddin, ASM and Chen, X},
  year={2018}
}

@inproceedings{foster2022drone,
  title={Drone footage wind turbine surface damage detection},
  author={Foster, Ashley and Best, Oscar and Gianni, Mario and Khan, Asiya and Collins, Keri and Sharma, Sanjay},
  booktitle={2022 IEEE 14th Image, Video, and Multidimensional Signal Processing Workshop (IVMSP)},
  pages={1--5},
  year={2022},
  organization={IEEE}
}

@article{chen2025weakly,
  title={Weakly-supervised semantic segmentation with image-level labels: from traditional models to foundation models},
  author={Chen, Zhaozheng and Sun, Qianru},
  journal={ACM Computing Surveys},
  volume={57},
  number={5},
  pages={1--29},
  year={2025},
  publisher={ACM New York, NY}
}

@inproceedings{yu2019uncertainty,
  title={Uncertainty-aware self-ensembling model for semi-supervised 3D left atrium segmentation},
  author={Yu, Lequan and Wang, Shujun and Li, Xiaomeng and Fu, Chi-Wing and Heng, Pheng-Ann},
  booktitle={International conference on medical image computing and computer-assisted intervention},
  pages={605--613},
  year={2019},
  organization={Springer}
}

@article{kendall2017uncertainties,
  title={What uncertainties do we need in bayesian deep learning for computer vision?},
  author={Kendall, Alex and Gal, Yarin},
  journal={Advances in neural information processing systems},
  volume={30},
  year={2017}
}

@article{loshchilov2017decoupled,
  title={Decoupled weight decay regularization},
  author={Loshchilov, Ilya and Hutter, Frank},
  journal={arXiv preprint arXiv:1711.05101},
  year={2017}
}

@article{zhao2023fast,
  title={Fast segment anything},
  author={Zhao, Xu and Ding, Wenchao and An, Yongqi and Du, Yinglong and Yu, Tao and Li, Min and Tang, Ming and Wang, Jinqiao},
  journal={arXiv preprint arXiv:2306.12156},
  year={2023}
}

@article{tarvainen2017mean,
  title={Mean teachers are better role models: Weight-averaged consistency targets improve semi-supervised deep learning results},
  author={Tarvainen, Antti and Valpola, Harri},
  journal={Advances in neural information processing systems},
  volume={30},
  year={2017}
}

@inproceedings{liu2016ssd,
  title={Ssd: Single shot multibox detector},
  author={Liu, Wei and Anguelov, Dragomir and Erhan, Dumitru and Szegedy, Christian and Reed, Scott and Fu, Cheng-Yang and Berg, Alexander C},
  booktitle={European conference on computer vision},
  pages={21--37},
  year={2016},
  organization={Springer}
}

@article{zhanfang2025enhancing,
  title={Enhancing wind turbine blade damage detection with YOLO-Wind},
  author={Zhanfang, Zhao and Tuo, Li},
  journal={Scientific Reports},
  volume={15},
  number={1},
  pages={18667},
  year={2025},
  publisher={Nature Publishing Group UK London}
}

@inproceedings{selvaraju2017grad,
  title={Grad-cam: Visual explanations from deep networks via gradient-based localization},
  author={Selvaraju, Ramprasaath R and Cogswell, Michael and Das, Abhishek and Vedantam, Ramakrishna and Parikh, Devi and Batra, Dhruv},
  booktitle={Proceedings of the IEEE international conference on computer vision},
  pages={618--626},
  year={2017}
}

@inproceedings{ahn2018learning,
  title={Learning pixel-level semantic affinity with image-level supervision for weakly supervised semantic segmentation},
  author={Ahn, Jiwoon and Kwak, Suha},
  booktitle={Proceedings of the IEEE conference on computer vision and pattern recognition},
  pages={4981--4990},
  year={2018}
}

@inproceedings{wang2020self,
  title={Self-supervised equivariant attention mechanism for weakly supervised semantic segmentation},
  author={Wang, Yude and Zhang, Jie and Kan, Meina and Shan, Shiguang and Chen, Xilin},
  booktitle={Proceedings of the IEEE/CVF conference on computer vision and pattern recognition},
  pages={12275--12284},
  year={2020}
}

@inproceedings{dai2015boxsup,
  title={Boxsup: Exploiting bounding boxes to supervise convolutional networks for semantic segmentation},
  author={Dai, Jifeng and He, Kaiming and Sun, Jian},
  booktitle={Proceedings of the IEEE international conference on computer vision},
  pages={1635--1643},
  year={2015}
}

@inproceedings{lin2014microsoft,
  title={Microsoft coco: Common objects in context},
  author={Lin, Tsung-Yi and Maire, Michael and Belongie, Serge and Hays, James and Perona, Pietro and Ramanan, Deva and Doll{\'a}r, Piotr and Zitnick, C Lawrence},
  booktitle={European conference on computer vision},
  pages={740--755},
  year={2014},
  organization={Springer}
}

@inproceedings{radford2021learning,
  title={Learning transferable visual models from natural language supervision},
  author={Radford, Alec and Kim, Jong Wook and Hallacy, Chris and Ramesh, Aditya and Goh, Gabriel and Agarwal, Sandhini and Sastry, Girish and Askell, Amanda and Mishkin, Pamela and Clark, Jack and others},
  booktitle={International conference on machine learning},
  pages={8748--8763},
  year={2021},
  organization={PmLR}
}

@inproceedings{he2022masked,
  title={Masked autoencoders are scalable vision learners},
  author={He, Kaiming and Chen, Xinlei and Xie, Saining and Li, Yanghao and Doll{\'a}r, Piotr and Girshick, Ross},
  booktitle={Proceedings of the IEEE/CVF conference on computer vision and pattern recognition},
  pages={16000--16009},
  year={2022}
}

@inproceedings{caron2021emerging,
  title={Emerging properties in self-supervised vision transformers},
  author={Caron, Mathilde and Touvron, Hugo and Misra, Ishan and J{\'e}gou, Herv{\'e} and Mairal, Julien and Bojanowski, Piotr and Joulin, Armand},
  booktitle={Proceedings of the IEEE/CVF international conference on computer vision},
  pages={9650--9660},
  year={2021}
}

@inproceedings{rong2023boundary,
  title={Boundary-enhanced co-training for weakly supervised semantic segmentation},
  author={Rong, Shenghai and Tu, Bohai and Wang, Zilei and Li, Junjie},
  booktitle={Proceedings of the IEEE/CVF conference on computer vision and pattern recognition},
  pages={19574--19584},
  year={2023}
}

@article{natarajan2013learning,
  title={Learning with noisy labels},
  author={Natarajan, Nagarajan and Dhillon, Inderjit S and Ravikumar, Pradeep K and Tewari, Ambuj},
  journal={Advances in neural information processing systems},
  volume={26},
  year={2013}
}

@inproceedings{patrini2017making,
  title={Making deep neural networks robust to label noise: A loss correction approach},
  author={Patrini, Giorgio and Rozza, Alessandro and Krishna Menon, Aditya and Nock, Richard and Qu, Lizhen},
  booktitle={Proceedings of the IEEE conference on computer vision and pattern recognition},
  pages={1944--1952},
  year={2017}
}

@article{han2018co,
  title={Co-teaching: Robust training of deep neural networks with extremely noisy labels},
  author={Han, Bo and Yao, Quanming and Yu, Xingrui and Niu, Gang and Xu, Miao and Hu, Weihua and Tsang, Ivor and Sugiyama, Masashi},
  journal={Advances in neural information processing systems},
  volume={31},
  year={2018}
}

@article{li2020dividemix,
  title={Dividemix: Learning with noisy labels as semi-supervised learning},
  author={Li, Junnan and Socher, Richard and Hoi, Steven CH},
  journal={arXiv preprint arXiv:2002.07394},
  year={2020}
}

@inproceedings{tanno2019learning,
  title={Learning from noisy labels by regularized estimation of annotator confusion},
  author={Tanno, Ryutaro and Saeedi, Ardavan and Sankaranarayanan, Swami and Alexander, Daniel C and Silberman, Nathan},
  booktitle={Proceedings of the IEEE/CVF conference on computer vision and pattern recognition},
  pages={11244--11253},
  year={2019}
}

@inproceedings{xia2020robust,
  title={Robust early-learning: Hindering the memorization of noisy labels},
  author={Xia, Xiaobo and Liu, Tongliang and Han, Bo and Gong, Chen and Wang, Nannan and Ge, Zongyuan and Chang, Yi},
  booktitle={International conference on learning representations},
  year={2020}
}

@article{polyak1992acceleration,
  title={Acceleration of stochastic approximation by averaging},
  author={Polyak, Boris T and Juditsky, Anatoli B},
  journal={SIAM journal on control and optimization},
  volume={30},
  number={4},
  pages={838--855},
  year={1992},
  publisher={SIAM}
}
